\documentclass[10pt,twocolumn,letterpaper]{article}

\usepackage{iccv}              

\usepackage[dvipsnames]{xcolor}

\usepackage[accsupp]{axessibility}  

\definecolor{iccvblue}{rgb}{0.21,0.49,0.74}
\usepackage[pagebackref,breaklinks,colorlinks,allcolors=iccvblue]{hyperref}

\def\paperID{10309} 
\def\confName{ICCV}
\def\confYear{2025}

\title{Active Membership Inference Test (aMINT):\\ Enhancing Model Auditability with Multi-Task Learning}

\author{Daniel DeAlcala, Aythami Morales, Julian Fierrez, Gonzalo Mancera,
Ruben Tolosana, Javier Ortega-Garcia \\
Biometrics and Data Pattern Analytics Lab, Universidad Autonoma de Madrid, Spain\\
{ \tt\small daniel.dealcala@uam.es} {\tt\small aythami.morales@uam.es} {\tt\small julian.fierrez@uam.es} 
}

\begin{document}
\maketitle
\begin{abstract}

Active Membership Inference Test (aMINT) is a method designed to detect whether given data were used during the training of machine learning models. In Active MINT, we propose a novel multitask learning process that involves training simultaneously two models: the original or Audited Model, and a secondary model, referred to as the MINT Model, responsible for identifying the data used for training the Audited Model. This novel multi-task learning approach has been designed to incorporate the auditability of the model as an optimization objective during the training process of neural networks. The proposed approach incorporates intermediate activation maps as inputs to the MINT layers, which are trained to enhance the detection of training data. We present results using a wide range of neural networks, from lighter architectures such as MobileNet to more complex ones such as Vision Transformers, evaluated in 5 public benchmarks. Our proposed Active MINT achieves over 80\% accuracy in detecting if given data was used for training, significantly outperforming previous approaches in the literature. Our aMINT and related methodological developments contribute to increasing transparency in AI models, facilitating stronger safeguards in AI deployments to achieve proper security, privacy, and copyright protection\footnote{Code available in \url{https://github.com/DanieldeAlcala/Membership-Inference-Test.git}}. 

\end{abstract}    
\section{Introduction}
\label{sec:intro}

The rapid evolution of Artificial Intelligence (AI) in recent years has motivated legal frameworks to safeguard citizens' rights. Institutions like the European Union (EU) have taken a proactive stance, proposing regulations that establish rules and responsibilities for AI developers, for example, the AI Act introduced in June 2024 \cite{madiega2021artificial}. These regulations aim to ensure lawful compliance, protect fundamental rights, and achieve transparent, fair, and reliable AI technology. In particular, among all regulations, the EU imposes companies the registration of the trained AI models in an EU-managed database, allowing for external audits and oversight. This trend in legislating AI deployments is not only observed at the EU level. Recently, in October 2024, the White House raised similar concerns through a memo that underscored AI as a matter of national security, advocating for stronger oversight and access to AI technologies to protect citizens \cite{USA}. This initiative emphasizes the need for stronger security measures, allowing supervision to prevent possible abuses of citizens' rights.

\begin{figure}[t!]
\centering
\includegraphics[width=0.99\linewidth]{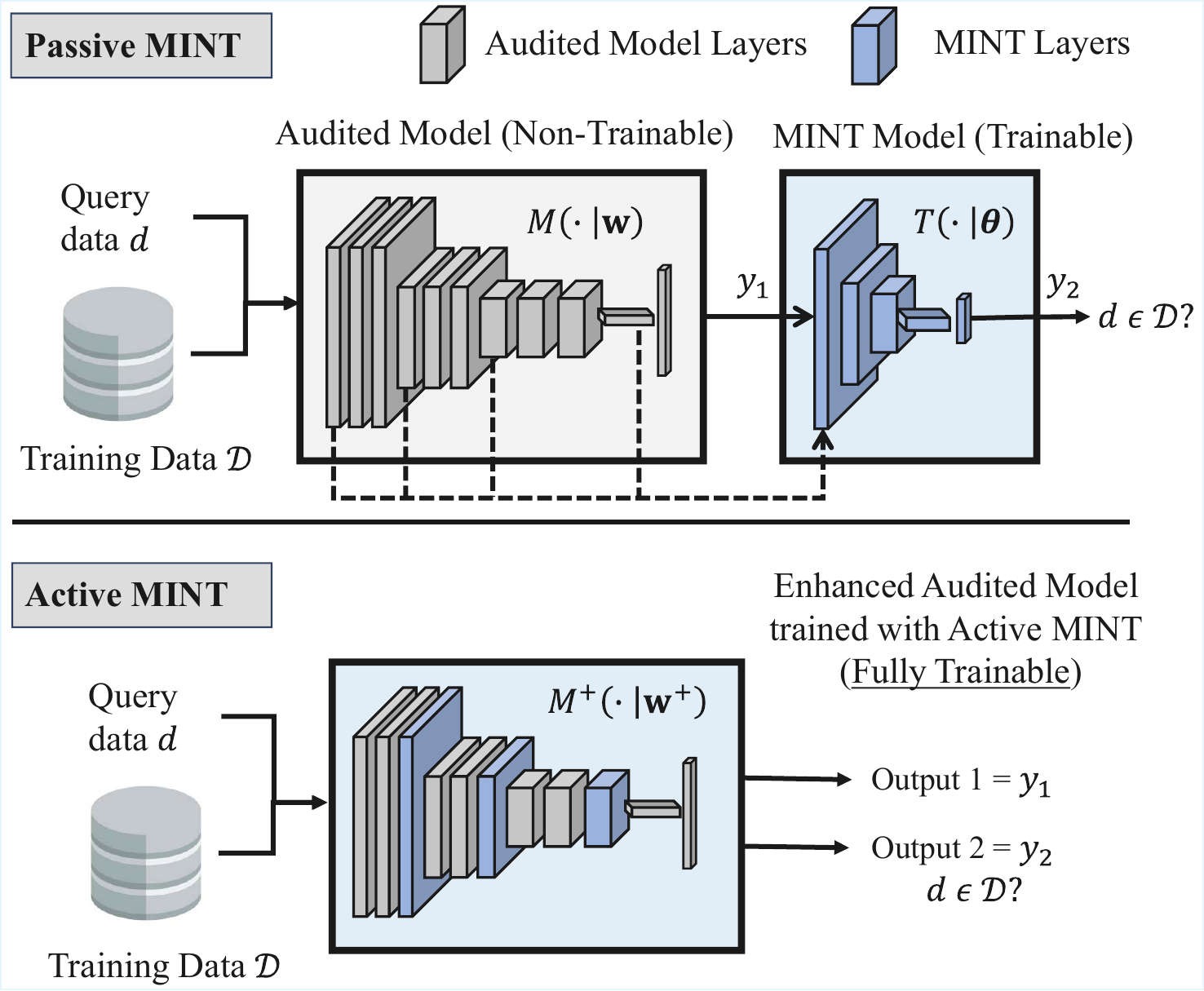} 
\caption{Differences between Passive MINT and Active MINT approaches. The Enhanced Audited model trained with the Active MINT approach ($M^{+}$) presents two outputs: 1) the primary task output $y$ (e.g., image classification label); 2) the MINT output (probability that data $d$ has been used during the training process of the model).}
\label{Block_Diagram_differences}
\end{figure}

The Membership Inference Test (MINT) \cite{dealcala2024my} emerged in 2024 as a research area focused on detecting whether specific data were used to train an AI model, with the aim of exposing unauthorized use of sensitive data, such as biometric information \cite{2021_TPAMI_SensitiveNets_Morales}. MINT builds on the field of Membership Inference Attacks (MIAs) \cite{shokri2017membership}, which attempt to extract private information (e.g. medical or consumer data) from trained models, revealing privacy vulnerabilities. These two fields (MIA and MINT) operate under different environmental conditions, resulting in differences in methodologies and outcomes. However, they share similarities, and research in one area can inspire progress in the other.

The main difference between these two lines of research, which leads to different environmental conditions, is that MIAs are attacks on the AI/ML model, meaning that collaboration from the model developer should not be considered. In contrast, MINT, as an auditing tool, allows for a certain level of collaboration with the model developer, such as limited access to the original model. This is supported by current and ongoing legislation, such as that of the White House \cite{USA} and the EU \cite{madiega2021artificial}, which impose model registration in accessible databases for authorized oversight. This is also supported by other legal frameworks such as the General Data Protection Regulation (GDPR) \cite{GDPR} and the California Consumer Privacy Act (CCPA) \cite{CCPA}.

Previous MINT research has assumed certain access to the model trained after it has been trained, focusing on analyzing patterns in the model's activations using the so-called MINT Model \cite{dealcala2024my, dealcala2024comprehensive,alcala25demo}. A similar approach is carried out in MIAs; however, since MIA is considered an attack, this is carried out on what are known as ``shadow models'', i.e. models that replicate the original model and over which complete control and access are available \cite{shokri2017membership,hu2022membership}. In this work, we explore a scenario where the developer actively engages in the development of the MINT Model, which we term Active MINT (see Fig. \ref{Block_Diagram_differences}). This approach contrasts with the previous studies where the model developer did not participate in the MINT Model's development, which we refer to as Passive MINT. In Active MINT, the owner of the model actively improves the auditability of the model by introducing MINT as an objective in his learning. 

The main contributions can be summarized as follows.

\begin{itemize}
    \item We propose Active MINT (aMINT), a novel multitask learning scheme in which two models are trained simultaneously: the Audited Model (based on neural network architectures) and a MINT Model that detects whether specific data were used during the Audited Model training.
    \item We present an extensive experimental section, evaluating models from lightweight architectures like MobileNet to more complex ones like ViT, across various image datasets, from simple digits to complex face images, demonstrating that aMINT improves auditability over existing MINT approaches.
    \item We compare our method with existing techniques, further extend their results, and develop new architectures. 
\end{itemize}

The article is structured as follows. Sect. \ref{sec:RelatedWorks} provides a review of the state of the art. Sect. \ref{sec:MINT} describes the main concepts of the proposed Active MINT and the specific details compared to the previous Passive MINT. The databases, AI models, and experimental protocols are described in Sect. \ref{sec:exp}. Results are discussed in Sec. \ref{sec:results}, whereas the discussion and conclusion are finally drawn in Sects. \ref{sec:Disc} and \ref{sec:Conclusion}. 

\section{Related Works}
\label{sec:RelatedWorks}

Membership Inference is a line of research introduced in 2017 by Shokri \etal \cite{shokri2017membership} under what they called Membership Inference Attacks (MIAs). This involved attempting to extract sensitive information that the model had been trained on. Later, in 2024, DeAlcala \etal introduced MINT \cite{dealcala2024my}, focusing on auditing models to ensure that they have been trained with legal data.

\subsection{Membership Inference Attacks (MIAs)}

Neural networks are often trained on sensitive or private data, posing a potential privacy risk \cite{privacy2025} to individuals if such information can be extracted in some way from trained models. MIAs were developed for this purpose, starting with Shokri \etal \cite{shokri2017membership}, who showed that training data can be extracted under specific conditions. Their approach involved training what they called ``Shadow Models'' that mimic the original model's behavior while granting full control over training data and unrestricted model access. Training these shadow models required detailed knowledge of the architecture of the original model, the training algorithm, and the dataset statistics. The statistics of the original dataset allowed them to replicate the dataset and use these dataset replicas to train shadow models with the same architecture and training algorithm. A binary classifier trained on the output classification vector of these shadow models then identifies data that are used in training or not. The underlying concept was that these shadow models would generalize to the original model, allowing the binary classifier to detect training data in the original model by extension.

Building on the work of Shokri \etal, the research line known as MIAs emerged, exploring various strategies for executing these attacks. Some studies, such as \cite{yeom2018privacy}, investigated applying a threshold directly to the loss values of individual data points rather than training a binary classifier, while others, such as \cite{song2021systematic,salem2018ml}, employed a threshold in the prediction output.

Based on the level of access to the model, Nasr \etal \cite{nasr2019comprehensive} defined two types of MIAs: Black-box, where only the model's output is available, and White-box, where internal information such as intermediate activations or gradients can be accessed. In their work, Nasr \etal introduced a White-box architecture that provided access to the model's intermediate activations and even gradients. They demonstrated that access to intermediate activations did not offer improvements over Black-box architectures. The best results were achieved using gradients. 

Another relevant study is that of Nasr \etal \cite{nasr2018machine}, which served as partial inspiration for our work. In their approach, the authors trained the original model to optimize its primary task while simultaneously regularizing its output to prevent MIAs from identifying training data.

In recent years, numerous studies have been presented that address the intrinsic complexity of the task \cite{shafran2021membership, rezaei2021difficulty}. These works also demonstrate that the results reported in the literature are often optimistic compared to reality because of insufficient experimental protocols. Furthermore, these concepts have been extended to other intriguing fields such as audio \cite{Shah2021,miao2019audio}, generative models \cite{hilprecht2019monte}, diffusion models \cite{duan2023diffusion}, and LLM/NLP \cite{mireshghallah2022quantifying, shejwalkar2021membership}.

\subsection{Membership Inference Test (MINT)}

Building on the ideas of MIAs, MINT introduces a new perspective \cite{dealcala2024my,alcala25demo}. While MIA is considered an attack, MINT is framed as an audit tool. This change in environmental conditions removes the need for shadow model training and allows the possibility of requesting information from the model developer. These differences in environmental conditions lead to different methods and results. The authors demonstrate that their ``MINT models'' can identify training data with high precision. They conducted experiments on facial recognition \cite{dealcala2024my} and general object classification models \cite{mancera25aaai}, exploring different MINT model architectures with different available data. Unlike MIAs, they show that White-box architectures in MINT significantly outperform Black-box architectures.

The same authors expanded on their research with another study that examines the factors influencing the detection performance of MINT Models. In particular, they highlight the options available to the model developer to positively or negatively affect the detection accuracy \cite{dealcala2024comprehensive}.

Subsequent work in MINT has demonstrated the benefits of considering gradients (gMINT) \cite{alcala25cvprw}, which are especially useful when auditing LLMs \cite{mancera25mint-text}.

\section{Active MINT: Concept and Architecture}
\label{sec:MINT}

The driving force behind AI is data, and today it is common to train models with large amounts of data \cite{aldoseri2023re}. However, access to these data is often restricted by licenses or data protection laws \cite{GDPR,CCPA}. Data owners have the right to decide how their data are used. A model developer might use data without the necessary permissions, and a tool is needed to detect such misuse \cite{USA,madiega2021artificial}. This is the role of MINT. 


In Fig. \ref{Block_Diagram_differences}, a diagram is presented that illustrates the functioning of Passive MINT is presented. The auditing entity (whether it be an international organization like the EU or the White House or a smaller private entity such as a company) gains access to the Audited Model after it has been trained. With this access, the auditor trains the so-called MINT Model, which is responsible for determining whether specific data was used or not for training. This approach is what we have referred to as Passive MINT (pMINT). For more information on pMINT, see the full paper \cite{dealcala2024my}.


In the present work, we will explore a different perspective. Instead of training the MINT Model a posteriori (Passive MINT), here we consider the training of the MINT Model concurrently with the Audited Model. We will refer to this as Active MINT (aMINT). In Figure \ref{Block_Diagram_differences}, we can see a schematic representation of this idea.

\subsection{Terminology}
We consider that the Audited Model $M$ is trained with data $\mathcal{D}$ for a specific task. For a sample $d$ that originates from training data $\mathcal{D}$ or external data $\mathcal{E}$ ($\mathcal{E} \notin \mathcal{D}$) the model $M$ generates an output $y_1 = M(d|\textbf{w})$ based on $d$ and the trained model parameters $\textbf{w}$. Intermediate outputs, known as Auxiliary Auditable Data ($\textrm{AAD} = N(d|\textbf{w}')$) can also be obtained from the input $d$ and a subset of parameters $\textbf{w}' \subset \textbf{w}$. The $\textrm{AAD}$ is used as input in the MINT Model that generates the output $y_2 = T(N(d|\textbf{w}')|\boldsymbol{\theta})$. This terminology applied to our use case is represented in Fig. \ref{Block_Diagram_CNNActive}.

\subsection{Active MINT: Multi-task Learning Approach}

\begin{figure*}[t]
\centering
\includegraphics[width=0.98\linewidth]{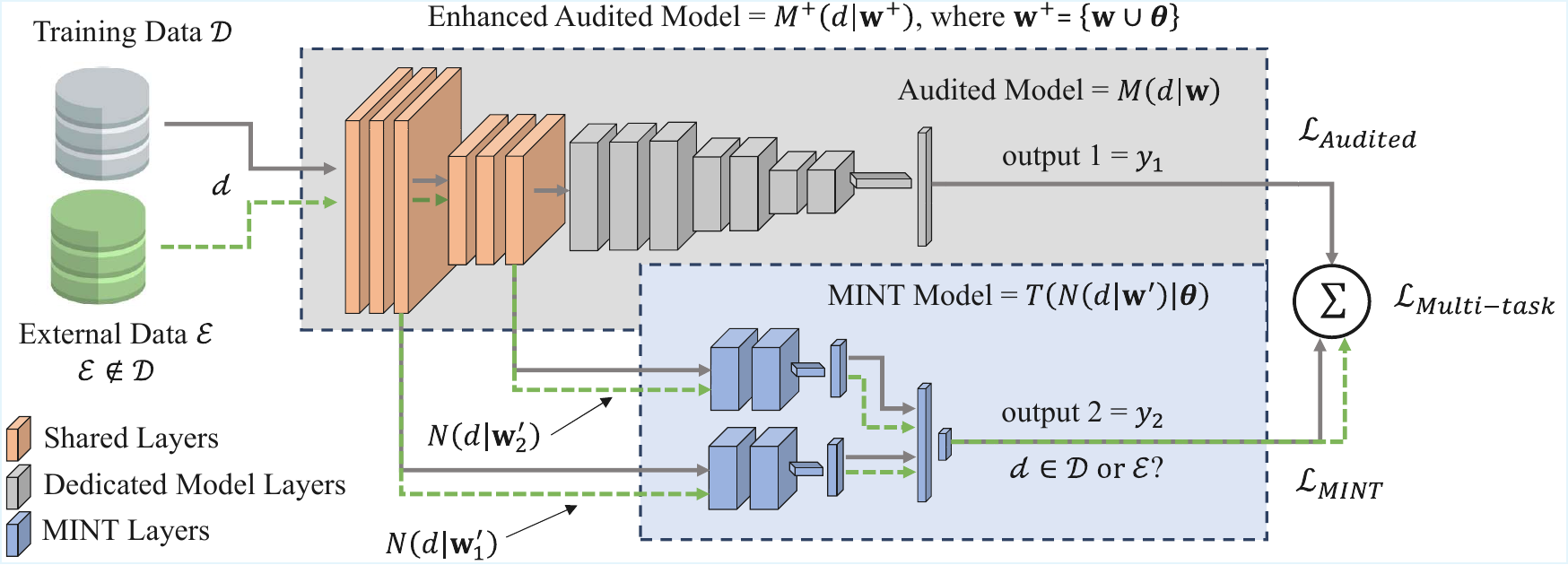} 
\caption{Full workflow of the Active MINT approach to train an Enhanced Audited Model ($M^+$), where the Audited Model ($M$) and the MINT Model ($T$) are jointly trained as a single multitask model that optimizes both tasks simultaneously.}
\label{Block_Diagram_CNNActive}
\end{figure*}

Fig. \ref{Block_Diagram_CNNActive} illustrates the full training workflow along with the implementation details. The goal of Active MINT is to create an Enhanced Audited Model $M^+$ that improves auditability, allowing to detect whether the input data $d$ belong to $\mathcal{D}$ or $\mathcal{E}$. This Enhanced Audited Model $M^+$ comprises the Audited Model $M$, trained using the dataset $\mathcal{D}$ (samples available to train Model $M$) for the audited task (e.g. image classification) and the MINT model $T$, which is trained on the $\textrm{AAD}$ for the MINT task. $M^+$ is defined by its parameters $\textbf{w}^{+}= \{\textbf{w} \cup \boldsymbol{\theta} \}$. The $\textrm{AAD}$ is generated by feeding both $\mathcal{D}$ and $\mathcal{E}$ (external data not available to train the Audited Model) through the model $M$.

In this work, $\textrm{AAD}$ is extracted from two points in the network, though extraction from one or more than two points is also possible. The model starts with shared layers, which then split at the point where the $\textrm{AAD}$ is extracted. The initial layers, shown in orange, are shared and need to be optimized for both the Audited Model task and the MINT task. Beyond this point, the remaining layers of model $M$ (in gray) are optimized solely for the Audited Model task, while the layers of model $T$ (in blue) are trained specifically for the MINT task. This shared optimization of the initial layers is a key distinction between Active MINT and Passive MINT, as Active MINT introduces common layers that are trained simultaneously for both tasks.

The batches contain samples from both datasets $\mathcal{D}$ and $\mathcal{E}$. The samples from $\mathcal{E}$ follow the path of the MINT Model (green lines in Fig. \ref{Block_Diagram_CNNActive}), while the samples from $\mathcal{D}$ traverse both the MINT Model path and the Audited Model path (gray lines in Fig. \ref{Block_Diagram_CNNActive}). Each path has its own loss function, and the losses from each path are combined to jointly train the model. For the model to train effectively, several considerations are essential when designing the multi-task loss function: both losses (the Audited Loss and the MINT Loss) must be normalized to remain within the same range, and a regularizer is included to balance the influence of each loss depending on the architecture and dataset:

\begin{equation}
\label{eqn:lossfunction}
\mathcal{L}_{\text{Multi-task}} = \lambda_1 \frac{\mathcal{L}_{\text{Audited}}}{\|\mathcal{L}_{\text{Audited}}\|} + \lambda_2 \frac{\mathcal{L}_{\text{MINT}}}{\|\mathcal{L}_{\text{MINT}}\|} + R(\textbf{w}^{+}),
\end{equation}

\noindent where $R(\textbf{w}^{+})$ is an L2 regularizer applied to the weights of the Enhanced Audited model, $\lambda_1$ and $\lambda_2$ are weighting factors to control the convergence of multitask learning, $\mathcal{L}_{\text{Audited}}$ is the loss function for the Audited Model (i.e., it depends on the model and the task for which the audited model is trained), and $\mathcal{L}_{\text{MINT}}$ is a binary cross-entropy loss. 




\subsection{Active MINT vs Passive MINT}

In both Active and Passive MINT, the MINT Model $T$ uses information on how the Audited Model $M$ processes the data $d$. In Passive MINT, the MINT Model $T$ is trained once the Audited Model $M$ has already been trained, requesting access to the developer only at that point. In contrast, the Active MINT approach proposed here requires the developer to train the MINT Model $T$ alongside their Audited Model $M$, thereby creating an Enhanced Audited Model $M^+$. It is crucial that this joint training does not significantly degrade the performance of the Audited Model $M$, as that would reduce its usefulness compared to Passive MINT strategies. In Sect. \ref{sec:exp}, we compare this original performance of the Audited Model with that of the Audited Model trained using the Active MINT approach.

Both approaches present advantages and drawbacks. In Passive MINT, the developer must allow access to the Audited model and provide some of the data used for training in order to train the MINT Model. In Active MINT, the developer does not need to disclose training data or grant model access, which can be crucial given the potential risks and the developers' preferences regarding sharing this information, even with regulatory bodies like the EU or US. However, unlike Passive MINT, the developer must actively participate in training the MINT Model.

\subsection{Active MINT: Deployment Considerations}
\label{sec:deployment}

Although Active MINT requires active developer participation, it does not mean that the auditor must fully trust them or lose control of the process. Several strategies can ensure that training is performed correctly and verifiably. One option is to provide the developer with a script that logs each training step (e.g., epochs, losses, and parameters) and digitally signs the logs using auditor-controlled keys. These logs could also be automatically uploaded to a remote server in real time for added transparency. Another strategy involves packaging the model, training code, and data into a closed Docker container, which can then be validated by comparing the hash of the final tarball to ensure that the exact intended setup was used. More advanced solutions, such as Multiparty Computation (MPC), can also be explored, for example, letting the developer train the Audited Model while the auditor retains control over the MINT Model.

Although the focus of this work is on presenting the core theoretical framework of aMINT, and not on deployment strategies, we believe it is important to briefly discuss these strategies to demonstrate the feasibility of applying aMINT in real-world, regulation-driven scenarios.

\section{Experimental Setup}
\label{sec:exp}

\subsection{Database and Models}

In Active MINT, unlike Passive MINT, the Audited Model must be trained alongside the MINT Model, creating the Enhanced Audited Model (Fig. \ref{Block_Diagram_CNNActive}). For our experiments, without loss of generality, we have selected to experiment with the task of Image Recognition as it offers a manageable environment for conducting a variety of experiments. Furthermore, most of the work on Membership Inference Attacks (MIAs) \cite{salem2018ml, yeom2018privacy, song2021systematic,ye2022enhanced,watson2021importance} has been conducted in this domain. Although MIAs and MINT differ in methodology and consequently in results, we have compared our results with recent MIA approaches in order to position our work in the state of the art. We present experiments with various architectures and datasets that we will mix together:

\begin{itemize} 
\item For the datasets, we have chosen: MNIST ($60$K training samples and $10$K test samples) \cite{lecun2010mnist}, CIFAR-10 ($50$K training samples and $10$K test samples) \cite{krizhevsky2009learning}, GTSRB ($39$K training samples and $13$K test samples) \cite{stallkamp2011german}, and finally Tiny Imagenet ($100$K training samples and $20$K test samples) \cite{chrabaszcz2017downsampled}. MNIST contains simple images of $10$ numbers with very low resolution ($28 \times 28$). This resolution slightly increases in CIFAR-10, with also $10$ classes where the realism grows significantly ($32 \times 32$). Next, we have GTSRB, which consists of fine-grained class data, differing more from the previous datasets and also featuring an increase in image resolution (from $15 \times 15$ the smallest to $222 \times 293$  the biggest) and number of classes ($43$ classes). Finally, Tiny Imagenet contains many more classes ($200$ classes) and images with higher resolution ($64 \times 64$). On the other hand, for the Face Recognition domain we have the CASIA WebFace dataset \cite{yi2014learning}, which consists of $500$K facial images ($250 \times 250$). 
\item Regarding architectures, we present experiments with MobileNet \cite{howard2017mobilenets}, ResNet50 \cite{he2016deep}, ResNet101 \cite{he2016deep}, DenseNet121 \cite{huang2017densely}, Xception \cite{chollet2017xception}, and ViT \cite{alexey2020image}.
\end{itemize}

\subsection{Experimental Protocol}
\label{sec:expprot}

In our experiments, we compare the proposed Active approach with existing Passive MINT approaches. The Audited Model and the general architecture of the MINT Model (which is optimized through parameter tuning, as outlined in \ref{sec:paramtuning}) will be the same in both approaches: Active MINT and Passive MINT. In the original work, the authors explored different forms of $\textrm{AAD}$ to train the MINT Models. Their primary finding was that using all activation maps as $\textrm{AAD}$, combined with a CNN-based MINT Model to analyze them, yielded the best results. They tested $\textrm{AAD}$ from various depths of the network, closer to both the input and output layers. Generally, they found that $\textrm{AAD}$ closer to the input yielded better results, although this varied slightly depending on the model.

In their paper, the authors did not explore the combination of $\textrm{AAD}$ from different depths. In our approach, we combine two activation maps as $\textrm{AAD}$ and analyze both using a CNN-based MINT Model, as shown in Fig. \ref{Block_Diagram_CNNActive}. We implement this combination in Passive MINT, which also means extending the method proposed by the authors \cite{dealcala2024my}. This allows the MINT Model to leverage more information and potentially achieve better results by selecting the most useful data for the MINT task. Although each model architecture (e.g., ResNet, DenseNet, etc.) is different, the overall process remains the same: selecting two activations maps and analyzing them with a CNN-based MINT Model. It is worth noting that for the ViT model, the activation maps generated are sequences, and thus, the MINT Model is comprised solely of Fully Connected layers.

Although in Passive MINT we know from DeAlcala \etal that activation maps closer to the model input generally perform better, in this new Active MINT approach, further experimentation is required to verify this behavior. Therefore, we present experiments using activation maps closer to the input layer (referred to as the Entry Setup), closer to the output layer (Output Setup), and finally, two intermediate activation maps (Middle Setup). This varies depending on the architecture, but to be more specific, these architectures are generally composed of different convolutional blocks, each containing multiple convolutional layers. In the Entry Setup, $\textrm{AAD}$ is extracted from the last two layers of the first convolutional block; in the Output Setup, it is obtained from the last two layers of the last convolutional block; and in the Middle Setup, it comes from the last layers of two intermediate convolutional blocks. 

The training subsets of the datasets used are divided into two parts: 50\% to train the original model, $\mathcal{D}$, and the other 50\% as external data, $\mathcal{E}$. This reduced training size explains the lower performance of $M$ compared to the state-of-the-art models trained on 100\% of the data. The performance of the Audited Model $M$ trained alone with this 50\% of the dataset is detailed in Sect.~\ref{sec:results}, allowing a direct comparison with the performance of the Audited Model in this Active MINT setup. All results in Sect.~\ref{sec:results} are calculated using the evaluation subsets of these datasets that guarantee no overlap between the training and evaluation data. 

\subsection{Hyperparameter Tuning}
\label{sec:paramtuning}

In both aMINT and pMINT architectures, various parameters require tuning to optimize the results. As mentioned, the MINT Models use a CNN architecture to analyze activation maps ($\textrm{AAD}$ in this work are two activation maps of size $H \times W \times C$), except in the case of ViT, where a fully connected network is used due to the sequential nature of its activation maps. The activation map sizes vary widely between experiments. Due to this diversity and the nature of multitask training, hyperparameters can vary significantly.

To illustrate, we compare two examples of MINT Model architectures: an entry setup using ResNet50 for MNIST (E1) and an entry setup using Xception for Tiny-Imagenet (E2). The dimensions of the activation map are $7\times7\times64$ and $7\times7\times128$ in E1, while in E2 they are both $16\times16\times728$.

In E1, due to the lower input dimensionality and the size of the dataset, the MINT architecture has fewer parameters to prevent overfitting, consisting of a single convolutional layer per path with $256$ channels, $3\times3$ filter size, and a stride of $1$. Then global grouping is applied and the two outputs are concatenated into a vector of size $512$, which is processed through two linear layers with a dropout value of $0.4$ in between. In E2, with higher dimensionality and a larger dataset, each path includes two convolutional layers, the first with $1024$ filters and the second with $2048$ both with $3\times3$ filter size and a stride of $1$. Global pooling is applied, producing a concatenated $4096$-dimensional vector, analyzed by two linear layers with a $0.2$ dropout.

Furthermore, the Audited Task of E1 is simpler than that of E2, so the ratio $\lambda_2\div\lambda_1$ (see Eq. \ref{eqn:lossfunction}) is set to $10$ for E1 and $10000$ for E2. For the learning rate (LR), E1 uses $10^{-5}$, while E2 uses $10^{-4}$, as ResNet requires a lower LR than Xception. Lastly, E1’s $R(\textbf{w}^{+})$ regularization term (see Eq. \ref{eqn:lossfunction}) is set to $10^{-4}$ to further mitigate overfitting, while E2 is set to $10^{-5}$. Both E1 and E2 followed an early stopping strategy, reaching around 50 epochs for E1 and 100 for E2.

\begin{table*}[]
\centering
\small
\begin{tabular}{l|llllllllllll|}
\cline{2-13}
                                                    & \multicolumn{12}{c|}{MNIST ($10$ classes)}                                                                                                                                                                   \\ \hline
\multicolumn{1}{|c|}{\multirow{2}{*}{Passive MINT}} & \multicolumn{2}{c|}{MobileNet}                               & \multicolumn{2}{c|}{ResNet50}                                & \multicolumn{2}{c|}{ResNet100}                               & \multicolumn{2}{c|}{DenseNet121}                             & \multicolumn{2}{c|}{Xception}                                & \multicolumn{2}{c|}{ViT}                              \\ \cline{2-13} 
\multicolumn{1}{|c|}{}                              & \multicolumn{1}{l|}{MINT } & \multicolumn{1}{l|}{Aud } & \multicolumn{1}{l|}{MINT } & \multicolumn{1}{l|}{Aud } & \multicolumn{1}{l|}{MINT } & \multicolumn{1}{l|}{Aud } & \multicolumn{1}{l|}{MINT } & \multicolumn{1}{l|}{Aud } & \multicolumn{1}{l|}{MINT } & \multicolumn{1}{l|}{Aud } & \multicolumn{1}{l|}{MINT } & Aud                \\ \hline
\multicolumn{1}{|l|}{Entry Setup}                  & \multicolumn{1}{c|}{0.86}         & \multicolumn{1}{c|}{0.96}        & \multicolumn{1}{c|}{0.83}         & \multicolumn{1}{c|}{0.97}        & \multicolumn{1}{c|}{0.83}         & \multicolumn{1}{c|}{0.97}        & \multicolumn{1}{c|}{0.82}         & \multicolumn{1}{c|}{0.99}        & \multicolumn{1}{c|}{0.85}         & \multicolumn{1}{c|}{0.98}        & \multicolumn{1}{c|}{0.80}         & \multicolumn{1}{c|}{0.94} \\ \hline                   
\multicolumn{1}{|l|}{Middle Setup}          & \multicolumn{1}{c|}{0.88}         & \multicolumn{1}{c|}{0.92}        & \multicolumn{1}{c|}{0.82}         & \multicolumn{1}{c|}{0.97}        & \multicolumn{1}{c|}{0.82}         & \multicolumn{1}{c|}{0.96}        & \multicolumn{1}{c|}{0.83}         & \multicolumn{1}{c|}{0.99}        & \multicolumn{1}{c|}{0.83}         & \multicolumn{1}{c|}{0.98}        & \multicolumn{1}{c|}{0.81}         & \multicolumn{1}{c|}{0.95} \\ \hline
\multicolumn{1}{|l|}{Output Setup}                  & \multicolumn{1}{c|}{0.82}         & \multicolumn{1}{c|}{0.90}        & \multicolumn{1}{c|}{0.80}         & \multicolumn{1}{c|}{0.90}        & \multicolumn{1}{c|}{0.77}         & \multicolumn{1}{c|}{0.80}        & \multicolumn{1}{c|}{0.81}         & \multicolumn{1}{c|}{0.98}        & \multicolumn{1}{c|}{0.80}         & \multicolumn{1}{c|}{0.83}        & \multicolumn{1}{c|}{0.80}         & \multicolumn{1}{c|}{0.93} \\ \hline
                                                    & \multicolumn{12}{c|}{CIFAR-10 ($10$ classes)}                                                                                                                                                         \\ \hline
\multicolumn{1}{|c|}{\multirow{2}{*}{Passive MINT}} & \multicolumn{2}{c|}{MobileNet}                               & \multicolumn{2}{c|}{ResNet50}                                & \multicolumn{2}{c|}{ResNet100}                               & \multicolumn{2}{c|}{DenseNet121}                             & \multicolumn{2}{c|}{Xception}                                & \multicolumn{2}{c|}{ViT}                              \\ \cline{2-13} 
\multicolumn{1}{|c|}{}                              & \multicolumn{1}{l|}{MINT } & \multicolumn{1}{l|}{Aud } & \multicolumn{1}{l|}{MINT } & \multicolumn{1}{l|}{Aud } & \multicolumn{1}{l|}{MINT } & \multicolumn{1}{l|}{Aud } & \multicolumn{1}{l|}{MINT } & \multicolumn{1}{l|}{Aud } & \multicolumn{1}{l|}{MINT } & \multicolumn{1}{l|}{Aud } & \multicolumn{1}{l|}{MINT } & Aud                \\ \hline
\multicolumn{1}{|l|}{Entry Setup}                  & \multicolumn{1}{c|}{0.86}         & \multicolumn{1}{c|}{0.41}        & \multicolumn{1}{c|}{0.86}         & \multicolumn{1}{c|}{0.53}        & \multicolumn{1}{c|}{0.87}         & \multicolumn{1}{c|}{0.58}        & \multicolumn{1}{c|}{0.86}         & \multicolumn{1}{c|}{0.80}        & \multicolumn{1}{c|}{0.86}         & \multicolumn{1}{c|}{0.64}        & \multicolumn{1}{c|}{0.86}         & \multicolumn{1}{c|}{0.19} \\ \hline                   
\multicolumn{1}{|l|}{Middle Setup}          & \multicolumn{1}{c|}{0.91}         & \multicolumn{1}{c|}{0.41}        & \multicolumn{1}{c|}{0.89}         & \multicolumn{1}{c|}{0.49}        & \multicolumn{1}{c|}{0.87}         & \multicolumn{1}{c|}{0.54}        & \multicolumn{1}{c|}{0.87}         & \multicolumn{1}{c|}{0.80}        & \multicolumn{1}{c|}{0.86}         & \multicolumn{1}{c|}{0.62}        & \multicolumn{1}{c|}{0.87}         & \multicolumn{1}{c|}{0.20} \\ \hline
\multicolumn{1}{|l|}{Output Setup}                  & \multicolumn{1}{c|}{0.85}         & \multicolumn{1}{c|}{0.33}        & \multicolumn{1}{c|}{0.85}         & \multicolumn{1}{c|}{0.39}        & \multicolumn{1}{c|}{0.85}         & \multicolumn{1}{c|}{0.40}        & \multicolumn{1}{c|}{0.82}         & \multicolumn{1}{c|}{0.76}        & \multicolumn{1}{c|}{0.87}         & \multicolumn{1}{c|}{0.46}        & \multicolumn{1}{c|}{0.88}         & \multicolumn{1}{c|}{0.19} \\ \hline

                                        & \multicolumn{12}{c|}{GSTRB ($43$ classes)}                                                                                                                                                                                                                                                                                                                                                      \\ \hline
\multicolumn{1}{|c|}{\multirow{2}{*}{Passive MINT}} & \multicolumn{2}{c|}{MobileNet}                               & \multicolumn{2}{c|}{ResNet50}                                & \multicolumn{2}{c|}{ResNet100}                               & \multicolumn{2}{c|}{DenseNet121}                             & \multicolumn{2}{c|}{Xception}                                & \multicolumn{2}{c|}{ViT}                              \\ \cline{2-13} 
\multicolumn{1}{|c|}{}                              & \multicolumn{1}{l|}{MINT } & \multicolumn{1}{l|}{Aud } & \multicolumn{1}{l|}{MINT } & \multicolumn{1}{l|}{Aud } & \multicolumn{1}{l|}{MINT } & \multicolumn{1}{l|}{Aud } & \multicolumn{1}{l|}{MINT } & \multicolumn{1}{l|}{Aud } & \multicolumn{1}{l|}{MINT } & \multicolumn{1}{l|}{Aud } & \multicolumn{1}{l|}{MINT } & Aud                \\ \hline
\multicolumn{1}{|l|}{Entry Setup}                  & \multicolumn{1}{c|}{0.89}         & \multicolumn{1}{c|}{0.89}        & \multicolumn{1}{c|}{0.86}         & \multicolumn{1}{c|}{0.98}        & \multicolumn{1}{c|}{0.87}         & \multicolumn{1}{c|}{0.97}        & \multicolumn{1}{c|}{0.85}         & \multicolumn{1}{c|}{0.99}        & \multicolumn{1}{c|}{0.86}         & \multicolumn{1}{c|}{0.99}        & \multicolumn{1}{c|}{0.80}         & \multicolumn{1}{c|}{0.91} \\ \hline                   
\multicolumn{1}{|l|}{Middle Setup}          & \multicolumn{1}{c|}{0.87}         & \multicolumn{1}{c|}{0.82}        & \multicolumn{1}{c|}{0.83}         & \multicolumn{1}{c|}{0.98}        & \multicolumn{1}{c|}{0.85}         & \multicolumn{1}{c|}{0.98}        & \multicolumn{1}{c|}{0.84}         & \multicolumn{1}{c|}{0.99}        & \multicolumn{1}{c|}{0.88}         & \multicolumn{1}{c|}{0.98}        & \multicolumn{1}{c|}{0.81}         & \multicolumn{1}{c|}{0.93} \\ \hline
\multicolumn{1}{|l|}{Output Setup}                  & \multicolumn{1}{c|}{0.83}         & \multicolumn{1}{c|}{0.81}        & \multicolumn{1}{c|}{0.81}         & \multicolumn{1}{c|}{0.95}        & \multicolumn{1}{c|}{0.81}         & \multicolumn{1}{c|}{0.94}        & \multicolumn{1}{c|}{0.83}         & \multicolumn{1}{c|}{0.99}        & \multicolumn{1}{c|}{0.88}         & \multicolumn{1}{c|}{0.95}        & \multicolumn{1}{c|}{0.79}         & \multicolumn{1}{c|}{0.90} \\ \hline
                                                    & \multicolumn{12}{c|}{Tiny ImageNet ($200$ classes)}                                                                                                                                                                                                                                                                                                                                              \\ \hline
\multicolumn{1}{|c|}{\multirow{2}{*}{Passive MINT}} & \multicolumn{2}{c|}{MobileNet}                               & \multicolumn{2}{c|}{ResNet50}                                & \multicolumn{2}{c|}{ResNet100}                               & \multicolumn{2}{c|}{DenseNet121}                             & \multicolumn{2}{c|}{Xception}                                & \multicolumn{2}{c|}{ViT}                              \\ \cline{2-13} 
\multicolumn{1}{|c|}{}                              & \multicolumn{1}{l|}{MINT } & \multicolumn{1}{l|}{Aud } & \multicolumn{1}{l|}{MINT } & \multicolumn{1}{l|}{Aud } & \multicolumn{1}{l|}{MINT } & \multicolumn{1}{l|}{Aud } & \multicolumn{1}{l|}{MINT } & \multicolumn{1}{l|}{Aud } & \multicolumn{1}{l|}{MINT } & \multicolumn{1}{l|}{Aud } & \multicolumn{1}{l|}{MINT } & Aud                \\ \hline
\multicolumn{1}{|l|}{Entry Setup}                  & \multicolumn{1}{c|}{0.80}         & \multicolumn{1}{c|}{0.11}        & \multicolumn{1}{c|}{0.80}         & \multicolumn{1}{c|}{0.12}        & \multicolumn{1}{c|}{0.80}         & \multicolumn{1}{c|}{0.11}        & \multicolumn{1}{c|}{0.86}         & \multicolumn{1}{c|}{0.34}        & \multicolumn{1}{c|}{0.88}         & \multicolumn{1}{c|}{0.28}        & \multicolumn{1}{c|}{0.81}         & \multicolumn{1}{c|}{0.17} \\ \hline                   
\multicolumn{1}{|l|}{Middle Setup}          & \multicolumn{1}{c|}{0.88}         & \multicolumn{1}{c|}{0.10}        & \multicolumn{1}{c|}{0.80}         & \multicolumn{1}{c|}{0.09}        & \multicolumn{1}{c|}{0.86}         & \multicolumn{1}{c|}{0.09}        & \multicolumn{1}{c|}{0.88}         & \multicolumn{1}{c|}{0.34}        & \multicolumn{1}{c|}{0.87}         & \multicolumn{1}{c|}{0.28}        & \multicolumn{1}{c|}{0.81}         & \multicolumn{1}{c|}{0.19} \\ \hline
\multicolumn{1}{|l|}{Output Setup}                  & \multicolumn{1}{c|}{0.79}         & \multicolumn{1}{c|}{0.12}        & \multicolumn{1}{c|}{0.79}         & \multicolumn{1}{c|}{0.07}        & \multicolumn{1}{c|}{0.79}         & \multicolumn{1}{c|}{0.08}        & \multicolumn{1}{c|}{0.79}         & \multicolumn{1}{c|}{0.33}        & \multicolumn{1}{c|}{0.86}         & \multicolumn{1}{c|}{0.16}        & \multicolumn{1}{c|}{0.80}         & \multicolumn{1}{c|}{0.17} \\ \hline
\end{tabular}%
\caption{Enhanced Audited Model performance in terms of MINT and Audited Model accuracies (MINT and Aud respectively) across the three setups (Entry Setup, Middle Setup, and Output Setup) using datasets that range from simpler to more complex patterns, alongside models varying from basic to advanced architectures. Perfect classification accuracy is represented by $1$.}
\label{tab:setupstable}
\end{table*}

\section{Results}
\label{sec:results}

The first experiment presented examines the Active MINT results in the three setups presented in \ref{sec:expprot}: Entry, Middle, and Output, as shown in Table \ref{tab:setupstable}. Active MINT trains the Audited Task and the MINT Task simultaneously (Fig. \ref{Block_Diagram_CNNActive}), making it especially relevant to compare the performance across all three setups for both tasks. This experiment is conducted on the Object Recognition datasets, which offer a controlled and extensive environment to explore different dataset complexities (from simpler MNIST datasets to more complex Tiny ImageNet). We include as results: 1) the MINT Accuracy (MINT in Tables \ref{tab:setupstable}, \ref{tab:Activevspassive}) which refers to the binary classification accuracy between training data ($\mathcal{D}$) and external data ($\mathcal{E}$); 2) the Audited Model Accuracy (Aud in Tables \ref{tab:setupstable}, \ref{tab:Activevspassive}) which refers to the performance of the model for the Audited Model task (e.g., image recognition as developed in our experiments). In Table \ref{tab:setupstable}, we can observe several noteworthy results:
\begin{itemize}
    \item MINT accuracy: Depending on the model and dataset, the Entry or Middle setups yield the best results. With MNIST, both setups achieve similar performance. In CIFAR-10 and Tiny ImageNet, the Middle setup achieves higher MINT accuracy, whereas in a more fine-grained dataset like GTSRB, the Entry setup slightly outperforms the others. However, key observation is that the Output setup does not provide better results in any dataset, with the only case where it matches the other setups being the Xception model, where all three setups show similar performance across datasets.
    \item Audited accuracy: The results follow a similar trend. Entry and Middle setups perform comparably, although in this case the Entry setup appears slightly advantageous, except for Tiny ImageNet. Again, the Output setup consistently yields the lowest performance.
\end{itemize}

The main takeaway is that the Output Setup offers no benefit. This outcome can be attributed to the fact that the Audited Task and the MINT Task are fundamentally opposed. The Audited Task aims to generalize well, ensuring similar performance on unseen samples, while the MINT Task works against this generalization, aiming to distinguish between samples used in training and those not. In the Output Setup, both tasks share a substantial portion of the layers (red layers in \ref{Block_Diagram_CNNActive}), which complicates the training since these layers must optimize for contradictory goals.

Therefore, in subsequent experiments where we compare this technology, we will use results from either of the other two setups (Entry Setup or Middle Setup). Specifically, we will proceed with the Entry Setup since, while both setups yield similar outcomes, it offers a slight advantage for the Audited Task. As discussed in Sect. \ref{sec:Disc}, minimizing performance loss on the Audited Task is essential for achieving a favorable trade-off for the developer. This advantage offered by the Entry setup regarding the Audited Accuracy is again due to the fact that the MINT Model and the Audited Model share fewer ``shared layers'' (red layers in Fig.~\ref{Block_Diagram_CNNActive}).

\begin{table*}[]
\centering
\small
\begin{tabular}{l|llllllllllll|}
\cline{2-13}
                                              & \multicolumn{12}{c|}{MNIST ($10$ classes)}                                                                                                                                                                                                                                                                                                                                                      \\ \hline
\multicolumn{1}{|c|}{\multirow{2}{*}{Method}} & \multicolumn{2}{c|}{MobileNet}                               & \multicolumn{2}{c|}{ResNet50}                                & \multicolumn{2}{c|}{ResNet100}                               & \multicolumn{2}{c|}{DenseNet121}                             & \multicolumn{2}{c|}{Xception}                                & \multicolumn{2}{c|}{ViT}                              \\ \cline{2-13} 
\multicolumn{1}{|c|}{}                        & \multicolumn{1}{l|}{MINT } & \multicolumn{1}{l|}{Aud } & \multicolumn{1}{l|}{MINT } & \multicolumn{1}{l|}{Aud } & \multicolumn{1}{l|}{MINT } & \multicolumn{1}{l|}{Aud } & \multicolumn{1}{l|}{MINT } & \multicolumn{1}{l|}{Aud } & \multicolumn{1}{l|}{MINT } & \multicolumn{1}{l|}{Aud } & \multicolumn{1}{l|}{MINT } & Aud                \\ \hline
\multicolumn{1}{|l|}{Passive MINT}            & \multicolumn{1}{c|}{0.52}         & \multicolumn{1}{c|}{0.97}        & \multicolumn{1}{c|}{0.51}         & \multicolumn{1}{c|}{0.97}        & \multicolumn{1}{c|}{0.51}         & \multicolumn{1}{c|}{0.98}        & \multicolumn{1}{c|}{0.52}         & \multicolumn{1}{c|}{0.99}        & \multicolumn{1}{c|}{0.54}         & \multicolumn{1}{c|}{0.99}        & \multicolumn{1}{c|}{0.53}         & \multicolumn{1}{c|}{0.97} \\ \hline
\multicolumn{1}{|l|}{Active MINT}                  & \multicolumn{1}{c|}{0.86}         & \multicolumn{1}{c|}{0.96}        & \multicolumn{1}{c|}{0.83}         & \multicolumn{1}{c|}{0.97}        & \multicolumn{1}{c|}{0.83}         & \multicolumn{1}{c|}{0.97}        & \multicolumn{1}{c|}{0.82}         & \multicolumn{1}{c|}{0.99}        & \multicolumn{1}{c|}{0.85}         & \multicolumn{1}{c|}{0.98}        & \multicolumn{1}{c|}{0.80}         & \multicolumn{1}{c|}{0.94} \\ \hline
                                                  & \multicolumn{12}{c|}{CIFAR-10 ($10$ classes)}                                                                                                                                                                                                                                                                                                                                                      \\ \hline
\multicolumn{1}{|c|}{\multirow{2}{*}{Method}} & \multicolumn{2}{c|}{MobileNet}                               & \multicolumn{2}{c|}{ResNet50}                                & \multicolumn{2}{c|}{ResNet100}                               & \multicolumn{2}{c|}{DenseNet121}                             & \multicolumn{2}{c|}{Xception}                                & \multicolumn{2}{c|}{ViT}                              \\ \cline{2-13} 
\multicolumn{1}{|c|}{}                        & \multicolumn{1}{l|}{MINT } & \multicolumn{1}{l|}{Aud } & \multicolumn{1}{l|}{MINT } & \multicolumn{1}{l|}{Aud } & \multicolumn{1}{l|}{MINT } & \multicolumn{1}{l|}{Aud } & \multicolumn{1}{l|}{MINT } & \multicolumn{1}{l|}{Aud } & \multicolumn{1}{l|}{MINT } & \multicolumn{1}{l|}{Aud } & \multicolumn{1}{l|}{MINT } & Aud                \\ \hline
\multicolumn{1}{|l|}{Passive MINT}            & \multicolumn{1}{c|}{0.66}         & \multicolumn{1}{c|}{0.41}        & \multicolumn{1}{c|}{0.66}         & \multicolumn{1}{c|}{0.55}        & \multicolumn{1}{c|}{0.69}         & \multicolumn{1}{c|}{0.59}        & \multicolumn{1}{c|}{0.60}         & \multicolumn{1}{c|}{0.80}        & \multicolumn{1}{c|}{0.67}         & \multicolumn{1}{c|}{0.66}        & \multicolumn{1}{c|}{0.60}         & \multicolumn{1}{c|}{0.20} \\ \hline
\multicolumn{1}{|l|}{Active MINT}                  & \multicolumn{1}{c|}{0.86}         & \multicolumn{1}{c|}{0.41}        & \multicolumn{1}{c|}{0.86}         & \multicolumn{1}{c|}{0.53}        & \multicolumn{1}{c|}{0.87}         & \multicolumn{1}{c|}{0.58}        & \multicolumn{1}{c|}{0.86}         & \multicolumn{1}{c|}{0.80}        & \multicolumn{1}{c|}{0.86}         & \multicolumn{1}{c|}{0.64}        & \multicolumn{1}{c|}{0.86}         & \multicolumn{1}{c|}{0.19} \\ \hline
                                              & \multicolumn{12}{c|}{GSTBR ($43$ classes)}                                                                                                                                                                                                                                                                                                                                                      \\ \hline
\multicolumn{1}{|c|}{\multirow{2}{*}{Method}} & \multicolumn{2}{c|}{MobileNet}                               & \multicolumn{2}{c|}{ResNet50}                                & \multicolumn{2}{c|}{ResNet100}                               & \multicolumn{2}{c|}{DenseNet121}                             & \multicolumn{2}{c|}{Xception}                                & \multicolumn{2}{c|}{ViT}                              \\ \cline{2-13} 
\multicolumn{1}{|c|}{}                        & \multicolumn{1}{l|}{MINT } & \multicolumn{1}{l|}{Aud } & \multicolumn{1}{l|}{MINT } & \multicolumn{1}{l|}{Aud } & \multicolumn{1}{l|}{MINT } & \multicolumn{1}{l|}{Aud } & \multicolumn{1}{l|}{MINT } & \multicolumn{1}{l|}{Aud } & \multicolumn{1}{l|}{MINT } & \multicolumn{1}{l|}{Aud } & \multicolumn{1}{l|}{MINT } & Aud                \\ \hline
\multicolumn{1}{|l|}{Passive MINT}            & \multicolumn{1}{c|}{0.59}         & \multicolumn{1}{c|}{0.92}        & \multicolumn{1}{c|}{0.61}         & \multicolumn{1}{c|}{0.98}        & \multicolumn{1}{c|}{0.61}         & \multicolumn{1}{c|}{0.98}        & \multicolumn{1}{c|}{0.61}         & \multicolumn{1}{c|}{0.99}        & \multicolumn{1}{c|}{0.59}         & \multicolumn{1}{c|}{0.99}        & \multicolumn{1}{c|}{0.60}         & \multicolumn{1}{c|}{0.96} \\ \hline
\multicolumn{1}{|l|}{Active MINT}             & \multicolumn{1}{c|}{0.89}         & \multicolumn{1}{c|}{0.89}        & \multicolumn{1}{c|}{0.86}         & \multicolumn{1}{c|}{0.98}        & \multicolumn{1}{c|}{0.87}         & \multicolumn{1}{c|}{0.97}        & \multicolumn{1}{c|}{0.85}         & \multicolumn{1}{c|}{0.99}        & \multicolumn{1}{c|}{0.86}         & \multicolumn{1}{c|}{0.99}        & \multicolumn{1}{c|}{0.80}         & \multicolumn{1}{c|}{0.91} \\ \hline
                                              & \multicolumn{12}{c|}{Tiny Imagenet ($200$ classes)}                                                                                                                                                                                                                                                                                                                                                      \\ \hline
\multicolumn{1}{|c|}{\multirow{2}{*}{Method}} & \multicolumn{2}{c|}{MobileNet}                               & \multicolumn{2}{c|}{ResNet50}                                & \multicolumn{2}{c|}{ResNet100}                               & \multicolumn{2}{c|}{DenseNet121}                             & \multicolumn{2}{c|}{Xception}                                & \multicolumn{2}{c|}{ViT}                              \\ \cline{2-13} 
\multicolumn{1}{|c|}{}                        & \multicolumn{1}{l|}{MINT } & \multicolumn{1}{l|}{Aud } & \multicolumn{1}{l|}{MINT } & \multicolumn{1}{l|}{Aud } & \multicolumn{1}{l|}{MINT } & \multicolumn{1}{l|}{Aud } & \multicolumn{1}{l|}{MINT } & \multicolumn{1}{l|}{Aud } & \multicolumn{1}{l|}{MINT } & \multicolumn{1}{l|}{Aud } & \multicolumn{1}{l|}{MINT } & Aud                \\ \hline
\multicolumn{1}{|l|}{Passive MINT}            & \multicolumn{1}{c|}{0.57}         & \multicolumn{1}{c|}{0.12}        & \multicolumn{1}{c|}{0.59}         & \multicolumn{1}{c|}{0.17}        & \multicolumn{1}{c|}{0.61}         & \multicolumn{1}{c|}{0.17}        & \multicolumn{1}{c|}{0.56}         & \multicolumn{1}{c|}{0.35}        & \multicolumn{1}{c|}{0.65}         & \multicolumn{1}{c|}{0.28}        & \multicolumn{1}{c|}{0.60}         & \multicolumn{1}{c|}{0.20} \\ \hline
\multicolumn{1}{|l|}{Active MINT}             & \multicolumn{1}{c|}{0.80}         & \multicolumn{1}{c|}{0.11}        & \multicolumn{1}{c|}{0.80}         & \multicolumn{1}{c|}{0.12}        & \multicolumn{1}{c|}{0.80}         & \multicolumn{1}{c|}{0.11}        & \multicolumn{1}{c|}{0.86}         & \multicolumn{1}{c|}{0.34}        & \multicolumn{1}{c|}{0.88}         & \multicolumn{1}{c|}{0.28}        & \multicolumn{1}{c|}{0.81}         & \multicolumn{1}{c|}{0.17} \\ \hline
                                              & \multicolumn{12}{c|}{CASIA Webface ($10,575$ classes)}                                                                                                                                                                                                                                                                                                                                                      \\ \hline
\multicolumn{1}{|c|}{\multirow{2}{*}{Method}} & \multicolumn{2}{c|}{MobileNet}                               & \multicolumn{2}{c|}{ResNet50}                                & \multicolumn{2}{c|}{ResNet100}                               & \multicolumn{2}{c|}{DenseNet121}                             & \multicolumn{2}{c|}{Xception}                                & \multicolumn{2}{c|}{ViT}                              \\ \cline{2-13} 
\multicolumn{1}{|c|}{}                        & \multicolumn{1}{l|}{MINT } & \multicolumn{1}{l|}{Aud } & \multicolumn{1}{l|}{MINT } & \multicolumn{1}{l|}{Aud } & \multicolumn{1}{l|}{MINT } & \multicolumn{1}{l|}{Aud } & \multicolumn{1}{l|}{MINT } & \multicolumn{1}{l|}{Aud } & \multicolumn{1}{l|}{MINT } & \multicolumn{1}{l|}{Aud } & \multicolumn{1}{l|}{MINT } & Aud                \\ \hline
\multicolumn{1}{|l|}{Passive MINT}            & \multicolumn{1}{c|}{0.60}         & \multicolumn{1}{c|}{0.17}        & \multicolumn{1}{c|}{0.61}         & \multicolumn{1}{c|}{0.12}        & \multicolumn{1}{c|}{0.61}         & \multicolumn{1}{c|}{0.14}        & \multicolumn{1}{c|}{0.62}         & \multicolumn{1}{c|}{0.31}        & \multicolumn{1}{c|}{0.63}         & \multicolumn{1}{c|}{0.17}        & \multicolumn{1}{c|}{0.59}         & \multicolumn{1}{c|}{0.10} \\ \hline
\multicolumn{1}{|l|}{Active MINT}             & \multicolumn{1}{c|}{0.86}         & \multicolumn{1}{c|}{0.15}        & \multicolumn{1}{c|}{0.80}         & \multicolumn{1}{c|}{0.11}        & \multicolumn{1}{c|}{0.81}         & \multicolumn{1}{c|}{0.13}        & \multicolumn{1}{c|}{0.82}         & \multicolumn{1}{c|}{0.29}        & \multicolumn{1}{c|}{0.84}         & \multicolumn{1}{c|}{0.17}        & \multicolumn{1}{c|}{0.76}         & \multicolumn{1}{c|}{0.09} \\ \hline

\end{tabular}%
\caption{Comparison between Active and Passive MINT in terms of MINT accuracy and Audited Model accuracy (MINT and Aud respectively). Perfect classification accuracy is represented by $1$.}
\label{tab:Activevspassive}
\end{table*}

We present a comparative analysis of Active and Passive MINT in Table~\ref{tab:Activevspassive}. As explained in Sect. \ref{sec:exp}, we expand the Passive MINT approach by using two Activation Maps as $\textrm{AAD}$, allowing for a consistent MINT Model architecture across Passive and Active MINT, with slight adjustments in parameter tuning for each model $M$ and dataset $D$. The activation maps in Passive MINT are also drawn from an entry-level setup, as this configuration yields the best results \cite{dealcala2024my}. The Audited Model also remains consistent across methods, ensuring that the comparison is as fair as possible.

In Active MINT, both models are trained jointly, requiring Face Recognition models to be trained from scratch with the MINT Model. It is important to note that the goal is not to develop competitive FR models but to compare Active and Passive MINT in this domain. FR accuracy is reported as the accuracy of identification in the $10,575$ identities used during training (random chance: $1/10575$).

In Table~\ref{tab:Activevspassive}, we present the results of the comparison between Active and Passive MINT for the MINT Model Task and the Audited Model Task. Additionally, in the Passive MINT rows, we indicate the performance of the Audited Task when trained independently, allowing us to evaluate any trade-offs in performance when training both tasks simultaneously. As shown, the MINT accuracy is considerably higher in Active MINT, surpassing Passive MINT across all models and datasets. However, this significant improvement in MINT accuracy comes with a drawback, as it slightly reduces the original Audited Task accuracy. As explained previously, this is due to the conflicting objectives of the MINT task, which oppose the generalization goal of the audit task. We discuss this trade-off in Sect.~\ref{sec:Disc}.

Due to the limited MINT literature, we only benchmark against \cite{dealcala2024comprehensive,alcala25demo}. In Table~\ref{tab:MIAMINT}, we also compare Active MINT with recent MIAs. As discussed previously, the environmental conditions between MIAs and MINT differ, since MIAs are considered attacks, whereas MINT functions as an auditing tool. Thus, unlike the comparison between Active and Passive MINT, this one occurs under different and often highly specific conditions used in MIA studies. However, this comparison serves to contextualize our results.

\begin{table}[]
\centering
\begin{tabular}{|l|cc|}
\hline
\multirow{2}{*}{Method} & \multicolumn{2}{c|}{ResNet50}                                    \\ \cline{2-3} 
                             & \multicolumn{1}{c|}{CIFAR-10} & \multicolumn{1}{c|}{GSTRB}       \\ \hline
Salem \textit{et al.} MIA \cite{salem2018ml}                 & \multicolumn{1}{c|}{0.61}         &           0.67               \\ \hline
Yeom \textit{et al.}  MIA \cite{yeom2018privacy}                  & \multicolumn{1}{c|}{0.64}         &           0.79               \\ \hline
Song \textit{et al.}  MIA \cite{song2021systematic}              & \multicolumn{1}{c|}{0.65}         &           0.68               \\ \hline
Ye \textit{et al.} MIA \cite{ye2022enhanced}                & \multicolumn{1}{c|}{0.52}         &           0.60               \\ \hline
Watson \textit{et al.} MIA \cite{watson2021importance}                & \multicolumn{1}{c|}{0.63}         &           0.79               \\ \hline
Passive MINT \cite{dealcala2024comprehensive,alcala25demo}                  & \multicolumn{1}{c|}{0.66} &    0.61             \\ \hline
Active MINT \textbf{(Ours)}                 & \multicolumn{1}{c|}{\textbf{0.86}}&    \textbf{0.86}             \\ \hline
\end{tabular}%
\caption{Comparison between recent works in MIA and our Active MINT using the same ResNet-50 architecture. Perfect classification accuracy is represented by $1$.}
\label{tab:MIAMINT}
\end{table}

\section{Discussion}
\label{sec:Disc}

The aim of MINT is to promote auditable AI aligned with emerging legislation and citizens' rights. Regulations in regions such as the EU and the US increasingly require transparent and accountable AI systems \cite{madiega2021artificial, USA}, requiring tools such as MINT. Achieving this goal requires the collaboration of developers, either voluntarily for public transparency or mandated by legislation. The objective of this work is to provide a tool that enables these goals, recognizing the legislative demand for such resources while avoiding involvement in legal matters.

The existing Passive MINT (pMINT) and the proposed Active MINT (aMINT) enable auditability by determining whether specific data were used for training. pMINT operates post-training and requires access to both the model and part of the training data. aMINT also requires developer collaboration, but does not need access to the model or any training data, thus avoiding the challenges of sharing sensitive information or granting access to private models. Active MINT on the other side requires the developer to train the MINT Model alongside their model. The objective is to make this process as simple as possible, aiming to keep it automated and transparent to the developer.

As shown in Sect.~\ref{sec:results}, aMINT achieves significantly higher accuracy than pMINT, making it a much more powerful tool for meeting fairness, transparency, and trustworthiness goals, with only a very small decrease in the Audited Model performance. This trade-off is highly favorable, even allowing some Audited Models to maintain their original performance. However, in certain cases, even a small drop in performance may not be acceptable. For these scenarios, it is crucial to ensure that Active MINT does not affect the performance of the Audited Model. As seen in Sect. \ref{sec:results}, this can be achieved by reducing the number of 'shared layers' (Fig. \ref{Block_Diagram_CNNActive}) between the MINT Model and the Audited Model. Another option would be to use pMINT methods, despite their lower MINT detection accuracy.


As future work, exploring advanced multi-task learning techniques could help further reduce performance degradation in high-stakes settings. Promising strategies include meta-optimization \cite{finn2017model}, conflict-averse training \cite{liu2021conflict}, and knowledge distillation \cite{li2020knowledge} to better balance both tasks and preserve model representations. Adapting these methods to aMINT is a promising direction for future research.

As future work, we also plan to connect MINT with related research in explainable AI \cite{2021_Computers_XAI_Ortega,2023_ECAIw_LFIT-XAI_Tello,ivan24gpt}, privacy- and bias-aware AI methods \cite{2020_AAAI_Discrimination_Serna,2023_COMPSAC_BiasAI_N-sigma_DeAlcala,peña2025bias,mancera2025pba}, and cryptographic constructions in pattern recognition \cite{2017_Access_HEmultiDTW_Marta,2017_PR_multiBtpHE_marta,2022_Access_DP-CL_Ahmad,pietro25ive,mahdi25block}.

\section{Conclusion}
\label{sec:Conclusion}

We proposed a method called Active MINT (aMINT), a novel auditing tool in line with the latest international regulation in trustworthy AI. aMINT enhances model auditability by training a MINT Model alongside the Audited Model (e.g., a classification model) to detect the data used in the training process. This results in an architecture composed of these two components, termed the Enhanced Audited Model. We present an extensive comparison between the proposed method and similar methods in the literature. Our results demonstrate the feasibility of Active MINT in a wide range of scenarios. Active MINT consistently detects the data used in training with a precision greater than 80\%, significantly outperforming previous approaches in the literature and opening a novel line of research aimed at improving the trustworthiness of AI models.
\section*{Acknowledgement}
\label{Acknowledgement}

BBforTAI (PID2021-127641OB-I00 MICINN/FEDER), HumanCAIC (TED2021-131787B-I00 MICINN), M2RAI (PID2024-160053OB-I00 MICIU/FEDER) and Cátedra ENIA UAM-Veridas (NextGenerationEU PRTR TSI-100927-2023-2). DeAlcala funded by FPU21/05785 and Mancera by PRE2022-104499. ELLIS~Unit~Madrid.
{
    \small
    \bibliographystyle{ieeenat_fullname}
    \bibliography{main}
}


\end{document}